\documentclass[11pt,a4paper]{article}

\usepackage[margin=1in]{geometry}
\usepackage{graphicx}
\usepackage{amsmath}
\usepackage{amssymb}
\usepackage{booktabs}
\usepackage{url}
\usepackage[colorlinks=true,linkcolor=blue,citecolor=blue,urlcolor=blue]{hyperref}
\usepackage[noblocks]{authblk}
\usepackage{tikz}
\usetikzlibrary{arrows.meta,positioning,shapes.geometric,calc}

\begin{document}

\title{A Multi-Agent Platform for Automated Enterprise Analytics and Insight Generation}

\author[1]{Manoj N M}
\author[1]{Vijayakrishna S}
\author[1]{Manjunath Srinivas}
\author[1]{Rohit Pahan}
\affil[1]{Rakuten India Enterprise Private Limited, Bengaluru, India \\
\texttt{\{manoj.m, vijayakrishna.s, manjunath.s, rohit.pahan\}@rakuten.com}}

\date{}

\maketitle

\begin{abstract}
This paper proposes a multi-agent framework built on CrewAI \cite{crewai} for conversational business intelligence. Five specialized AI agents operate in a sequential pipeline to process natural language queries, retrieve and analyze data, generate visualizations via the Model Context Protocol (MCP) \cite{mcp}, and deliver actionable insights. The platform features a defense-in-depth security architecture for multi-tenant data isolation and a query parameterization mechanism for transforming conversational insights into reusable dashboard components. Evaluation across 300 end-to-end test cases spanning synthetic and production enterprise datasets demonstrates 95.3\% functional accuracy, a mean response latency of 24 seconds, and a response quality score of 4.52/5.0 as assessed by an LLM-as-a-Judge framework, with a 93.0\% hallucination-free rate, representing a 22.6 percentage point accuracy improvement and 20.2\% quality gain over a single-agent baseline. Cross-model evaluation across four LLM backends and human expert validation confirm architectural generalizability and evaluator reliability. An ablation study confirms that the Data Analysis and Report Aggregation agents are the primary drivers of output quality.
\end{abstract}


\section{Introduction}
Modern enterprises generate vast amounts of data distributed across heterogeneous storage systems. Traditional analysis methods require manual SQL formulation or complex business intelligence tools such as Looker Studio or Tableau, making actionable insight generation a time-consuming and labor-intensive process.

We propose a conversational multi-agent platform that automates data interaction by decomposing complex user queries into discrete steps (planning, data retrieval, analysis, and insight aggregation), each handled by a specialized agent with domain-specific tools. We additionally introduce a standardized evaluation framework combining automated functional testing with LLM-based quality assessment to validate the reliability of multi-agentic analytical solutions.

\section{Literature Review}
Recent advancements in Large Language Models (LLMs) have driven a shift from single-agent analytical tools to dynamic multi-agent systems (MAS) capable of complex reasoning and autonomous task execution. We organize related work into two thematic areas.

\subsection{Multi-Agent Systems and Conversational BI}
Yang et al. \cite{yang2024lamas} categorize LLM-based Multi-Agent System architectures into Star, Ring, Graph, and Bus patterns, emphasizing standardized protocols for collaboration. Hong et al. \cite{hong2023metagpt} propose MetaGPT, encoding standardized operating procedures into agent prompts with role differentiation, reducing hallucinations through structured workflows. Zhang et al. \cite{zhang2025agentorchestra} introduce AgentOrchestra, a hierarchical planning framework using the Tool-Environment-Agent (TEA) protocol that outperforms flat-agent baselines. Becker \cite{becker2024multiagent} empirically finds that multi-agent systems with expert personas excel at complex reasoning but risk task ``drift'' in free-form conversation, a pitfall our sequential pipeline mitigates through structured workflow enforcement.

For enterprise deployment, Wang et al. \cite{wang2025enterprise} propose a four-layer framework addressing text-to-SQL challenges through RAG and a dual-layer security architecture. Jiang et al. \cite{jiang2025siriusbi} present SiriusBI, achieving 93--96\% SQL accuracy through multi-round dialogue refinement deployed across Tencent's business sectors. Ni et al. \cite{ni2025intellidispatch} enhance Text-to-SQL reliability through chain-of-thought reasoning \cite{wei2022chain} and SQL validation modules for the power systems domain.

\subsection{LLM-Driven Analytics and Visualization}
Zhang and Elhamod \cite{zhang2025data2dashboard} present a Data-to-Dashboard framework using modular LLM agents with iterative self-reflection to generate domain-grounded insights and semantically meaningful visualizations. Bai et al. \cite{bai2025insight} detail Insight Agents, a deployed hierarchical manager-worker system for e-commerce that balances accuracy with low-latency requirements through a hybrid approach combining powerful LLMs with lightweight routing models.

For visualization, Wang et al. \cite{wang2023llm4vis} introduce LLM4Vis, a ChatGPT-based chart recommendation system that matches or outperforms traditional ML recommenders via in-context learning. Goswami et al. \cite{goswami2025plotgen} present PlotGen, automating scientific plotting through multi-agent collaboration with multimodal feedback loops for iterative refinement. Our system extends beyond recommendation to full chart generation, additionally managing data access, query parameterization, and multi-tenant security.

\section{System Architecture}

\subsection{Architectural Overview}
\begin{figure}[t]
  \centering
  \includegraphics[width=\textwidth,keepaspectratio]{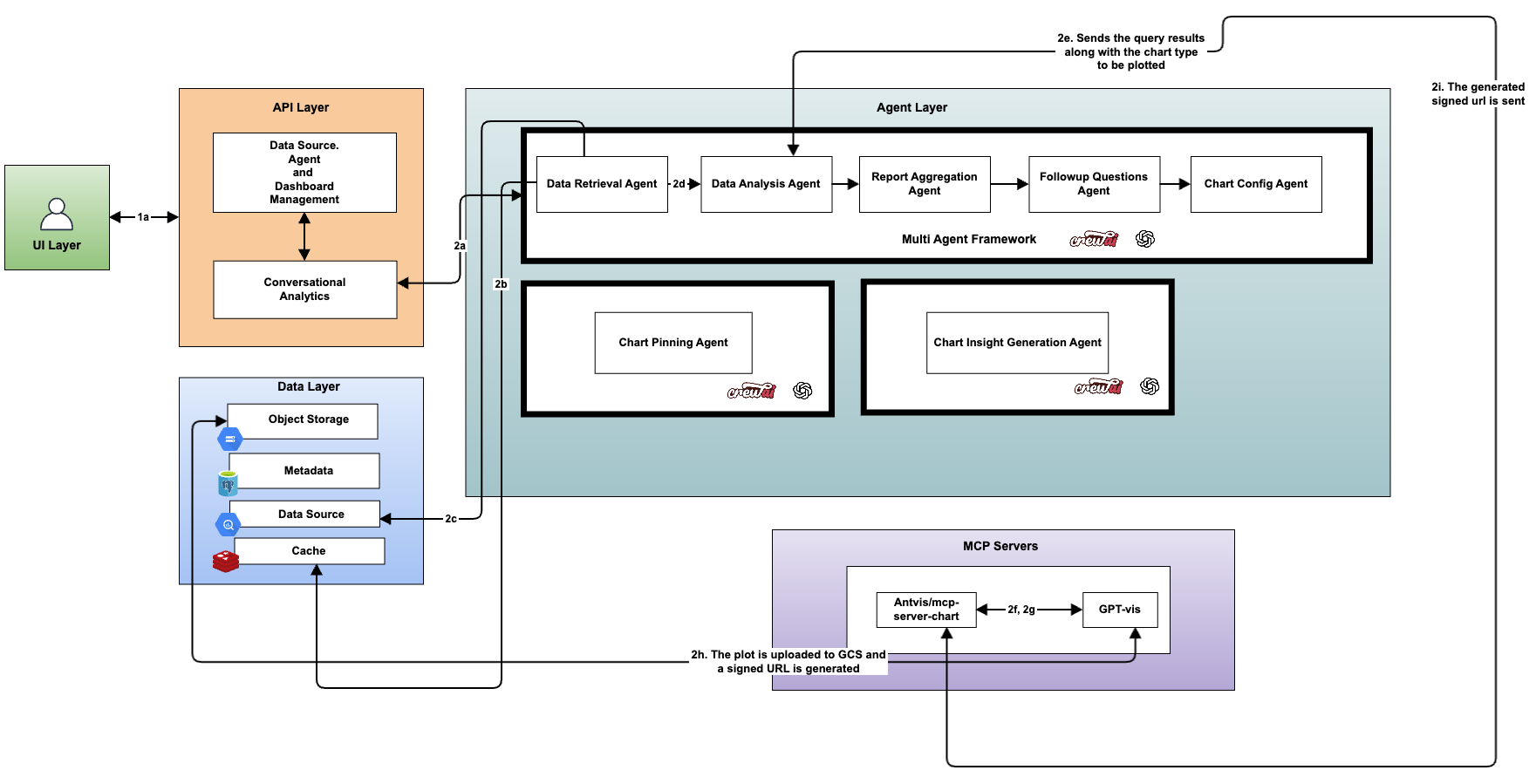}
  \caption{High-Level System Architecture}
  \label{fig:architecture}
\end{figure}

Our platform implements a layered microservices architecture comprising five tiers (Figure~\ref{fig:architecture}): a \textbf{UI Layer} for conversational interaction; an \textbf{API Layer} handling data source management, user sessions, and query routing; an \textbf{Agent Layer} powered by CrewAI \cite{crewai} for multi-agent orchestration; a \textbf{Data Layer} providing object storage, metadata stores, data warehouse connections, and caching; and an \textbf{MCP Layer} \cite{mcp} enabling visualization through Antvis/mcp-server-chart integration.

\subsection{Agent Crew Composition}
The core of the system is a five-agent sequential crew:

\begin{itemize}
    \item \textbf{Data Retrieval Agent:} Maps natural language queries to data sources, generates executable SQL with schema awareness, and enforces read-only access against authorized tables.
    \item \textbf{Data Analysis Agent:} Interprets query results to identify patterns, trends, and anomalies; generates charts via MCP based on data characteristics.
    \item \textbf{Report Aggregation Agent:} Synthesizes prior outputs into cohesive markdown responses with language detection and translation for multilingual deployments.
    \item \textbf{Follow-up Questions Agent:} Generates contextual follow-up suggestions in conversational language to guide deeper exploration.
    \item \textbf{Chart Configuration Agent:} Produces visualization metadata (chart types, axis mappings, labels) for interactive UI rendering.
\end{itemize}

\subsection{Query Parameterization}
A distinguishing feature is the Query Parameterization capability that transforms ad-hoc analytical queries into reusable, dashboard-ready components. The system extracts literal values from WHERE clauses into parameterized filters, leverages schema introspection (partition columns, clustering fields) for intelligent parameterization decisions, performs dry-run cost estimation with materialization recommendations for expensive queries, and generates drill-down dimensions from SELECT and GROUP BY columns for interactive filtering.

\subsection{Model Context Protocol Integration}
The platform leverages MCP \cite{mcp} through a server-adapter pattern where the MCP server runs as an independent microservice exposing chart generation tools (bar, line, pie, scatter, funnel, treemap, sankey, and others). Agents discover available tools and schemas at runtime, enabling dynamic capability extension. Visualization tools generate charts and return cloud storage URLs directly, simplifying the asset pipeline.

\section{Methodology}

\subsection{Architectural Paradigm Selection}
We considered three candidate architectures. A \textbf{monolithic single-agent} approach suffers from cognitive overload and poor error isolation as the model simultaneously optimizes competing objectives \cite{becker2024multiagent}. A \textbf{hierarchical manager-worker} architecture \cite{bai2025insight} introduces unpredictability and communication overhead. We adopted a \textbf{sequential multi-agent pipeline} where specialized agents execute in predefined order, motivated by separation of concerns, predictable execution flow, independent testability, and graceful degradation.

\subsection{Agent Coordination}
\begin{figure}[h]
\centering
\begin{tikzpicture}[
    node distance=0.55cm,
    agentbox/.style={
        rectangle, draw, rounded corners=2pt, minimum width=4.8cm,
        minimum height=0.65cm, align=center, font=\small\bfseries,
        fill=blue!8, draw=blue!40
    },
    artifact/.style={
        font=\scriptsize\itshape, text=black!70, align=center
    },
    arr/.style={-{Stealth[length=2.5mm]}, thick, draw=blue!50},
    phasebox/.style={
        rectangle, draw=gray!40, dashed, rounded corners=2pt,
        minimum width=2.0cm, minimum height=0.55cm, align=center,
        font=\scriptsize, fill=gray!5
    }
]

% Nodes
\node[phasebox] (init) {Initialization \& Validation};
\node[agentbox, below=of init] (retrieval) {Data Retrieval Agent};
\node[agentbox, below=of retrieval] (analysis) {Data Analysis Agent};
\node[font=\scriptsize, text=blue!60, right=0.15cm of analysis, anchor=west] {(via MCP)};
\node[agentbox, below=of analysis] (aggregation) {Report Aggregation Agent};
\node[agentbox, below=of aggregation] (followup) {Follow-up Questions Agent};
\node[agentbox, below=of followup] (chart) {Chart Configuration Agent};
\node[phasebox, below=of chart] (post) {Post-processing \& Assembly};

% Arrows with artifact labels
\draw[arr] (init) -- node[artifact, right, xshift=2pt] {Sanitized query, schema, context} (retrieval);
\draw[arr] (retrieval) -- node[artifact, right, xshift=2pt] {SQL query + cached result set} (analysis);
\draw[arr] (analysis) -- node[artifact, right, xshift=2pt] {Patterns, insights, chart objects} (aggregation);
\draw[arr] (aggregation) -- node[artifact, right, xshift=2pt] {Formatted markdown report} (followup);
\draw[arr] (followup) -- node[artifact, right, xshift=2pt] {Contextual follow-up suggestions} (chart);
\draw[arr] (chart) -- node[artifact, right, xshift=2pt] {UI visualization metadata} (post);

% Input/output labels
\node[above=0.35cm of init, font=\small\bfseries] (input) {User Natural Language Query};
\draw[arr] (input) -- (init);
\node[below=0.35cm of post, font=\small\bfseries] (output) {Structured Analytical Response};
\draw[arr] (post) -- (output);

\end{tikzpicture}
\caption{Sequential Pipeline Execution Flow. Each agent receives cumulative outputs of preceding agents via context injection. Dashed boxes represent non-agent processing phases; solid boxes represent autonomous LLM agent stages.}
\label{fig:pipeline}
\end{figure}
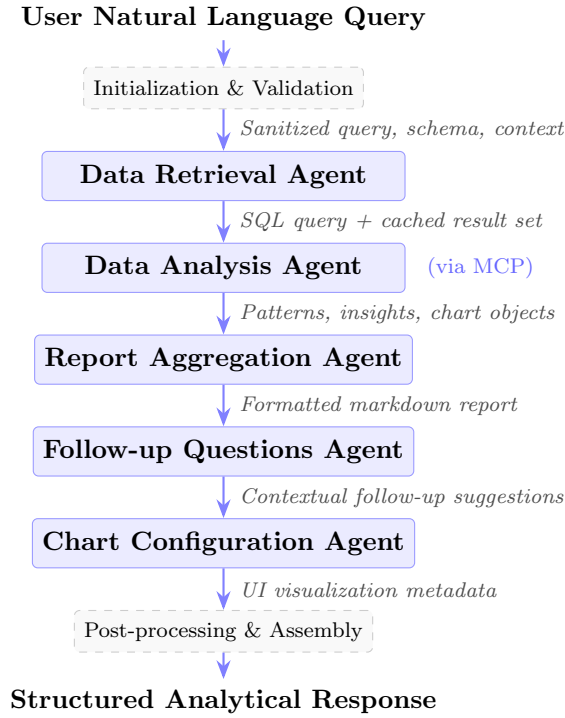

The framework employs CrewAI's \cite{crewai} sequential process mode. Execution proceeds through: (1) initialization with input validation and credential verification; (2) data retrieval with SQL generation and execution; (3) data analysis with pattern extraction and chart generation via MCP; (4) report aggregation with language translation; (5) follow-up question generation; (6) chart configuration; and (7) post-processing that restores cached query results and assembles the final response. Figure~\ref{fig:pipeline} illustrates this flow.

\subsection{Security Architecture}
We implement a defense-in-depth strategy with four layers: (1) \textbf{Input Validation} with query sanitization against SQL injection patterns and length limits; (2) \textbf{Query Guardrails} blocking unsafe operations (DROP, DELETE, INSERT, ALTER, CREATE, TRUNCATE) and enforcing table-level access control; (3) \textbf{Multi-tenant Isolation} through service account impersonation and tenant-specific credentials; (4) \textbf{Output Validation} with sensitive data scanning and error message sanitization.

\subsection{Prompt Engineering}
Agent prompts were developed through iterative refinement: role/goal definition, guardrail and anti-hallucination rule integration, edge case handling, and language/formatting specification.

\section{Evaluation}

We evaluate our platform across three dimensions: functional accuracy, response latency, and output quality, using 300 test cases assessed via automated functional testing, an LLM-as-a-Judge quality framework, and human expert validation.

\subsection{Experimental Setup}

\subsubsection{Test Dataset and Suite Design}
The evaluation employs two complementary datasets on Google BigQuery. The first is a synthetic e-commerce dataset with four relational tables (\texttt{customers}, \texttt{orders}, \texttt{order\_items}, \texttt{products}) totaling 12,500 records across four regions and five product categories. The second is a production competitive pricing intelligence dataset with four relational tables spanning two e-commerce platforms: two item catalog tables covering approximately 569,000 overlapping products with attributes including titles, multi-level category taxonomies, brand metadata, and pricing across up to 68 columns, and two price history tables containing approximately 1.17 million distinct price records capturing temporal pricing dynamics across 8.8 billion cumulative observations. The enterprise dataset contains Japanese-language product content, introducing multilingual complexity absent from the synthetic data.

We designed 300 test cases across eight categories: Simple Aggregations (65), Filtered Aggregations (35), Time Series Analysis (40), Multi-Table JOINs (35), Complex Analytics (30), Edge Cases (30), Multi-Turn Conversations (25), and Security Guardrails (40). Each analytical test validates SQL generation, execution correctness, result retrieval, and structured insight presence. Of the 260 analytical test cases, 100 are independently scored by human domain experts.

\subsubsection{LLM-as-a-Judge Methodology}
We employ an LLM-as-a-Judge framework \cite{zheng2023judging} using Claude Sonnet 4.5 as the evaluator. Each of the 260 analytical responses is assessed across five dimensions on a 1--5 Likert scale: \textit{Factual Accuracy}, \textit{Relevance}, \textit{Completeness}, \textit{Actionability}, and \textit{Language Quality}. The judge additionally performs hallucination detection by cross-referencing factual claims against raw query results. To mitigate evaluator bias, we use structured rubrics with explicit per-dimension scoring criteria and require grounding of every factual accuracy assessment in the raw data.

\subsubsection{Baseline, Ablation, and Cross-Model Configuration}
The single-agent baseline employs the same GPT-4.1 model and BigQuery tooling but consolidates all responsibilities into a single prompt, omitting the regex-based input validation layer. We conduct an ablation study selectively removing individual agents to quantify marginal contributions. To assess architectural generalizability, we replicate the full 300-test evaluation across four LLM backends: GPT-4.1, Claude Sonnet 4, Gemini 2.5 Flash, and Llama 3.1 70B (self-hosted). Three domain experts independently score 100 randomly sampled analytical responses using the same five-dimension rubric to validate the LLM-as-a-Judge methodology against human assessment.

\subsection{Results}

\subsubsection{End-to-End Functional Accuracy}
The system achieves 95.3\% accuracy (286/300 tests), including blocking all 40 adversarial inputs (Table~\ref{tab:accuracy}). Failures concentrate in complex analytics (deeply nested CTEs producing semantically incorrect SQL over the enterprise schema) and multi-turn conversations (cross-turn comparisons exceeding the reasoning window with large result sets).

\begin{table}[h]
\centering
\caption{End-to-End Functional Accuracy by Category}
\label{tab:accuracy}
\begin{tabular}{|l|c|c|c|}
\hline
\textbf{Category} & \textbf{Tests} & \textbf{Passed} & \textbf{Pass Rate} \\
\hline
Simple Aggregations & 65 & 64 & 98.5\% \\
Filtered Aggregations & 35 & 35 & 100\% \\
Time Series Analysis & 40 & 39 & 97.5\% \\
Multi-Table JOINs & 35 & 33 & 94.3\% \\
Complex Analytics & 30 & 26 & 86.7\% \\
Edge Cases & 30 & 28 & 93.3\% \\
Multi-Turn Conversations & 25 & 21 & 84.0\% \\
Security Guardrails & 40 & 40 & 100\% \\
\hline
\textbf{Total} & \textbf{300} & \textbf{286} & \textbf{95.3\%} \\
\hline
\end{tabular}
\end{table}

\subsubsection{Latency Analysis}
The overall mean latency of 23.8 seconds falls within the 120-second processing target (Table~\ref{tab:latency}). Complex analytics queries average 38.6s due to window functions and CTEs over the enterprise schema, while multi-turn queries exhibit low latency (mean 16.2s) by leveraging prior context.

\begin{table}[h]
\centering
\caption{End-to-End Latency by Query Category (seconds)}
\label{tab:latency}
\begin{tabular}{|l|c|c|c|c|}
\hline
\textbf{Category} & \textbf{$n$} & \textbf{Mean} & \textbf{Min} & \textbf{Max} \\
\hline
Simple Aggregations & 65 & 20.8 & 8.2 & 68.3 \\
Filtered Aggregations & 35 & 16.2 & 9.1 & 24.5 \\
Time Series Analysis & 40 & 19.4 & 11.8 & 38.2 \\
Multi-Table JOINs & 35 & 26.8 & 13.2 & 52.1 \\
Complex Analytics & 30 & 38.6 & 16.4 & 118.5 \\
Edge Cases & 30 & 28.1 & 8.4 & 65.8 \\
Multi-Turn Conversations & 25 & 16.2 & 10.8 & 22.4 \\
\hline
\textbf{Overall} & \textbf{260} & \textbf{23.8} & \textbf{8.2} & \textbf{118.5} \\
\hline
\end{tabular}
\end{table}

\subsubsection{Security Guardrail Evaluation}
All 40 adversarial attack vectors were intercepted (Table~\ref{tab:security}), including 34 additional vectors targeting the enterprise schema. The input validation layer handles known harmful patterns with sub-millisecond response, while the LLM guardrail layer catches sophisticated attacks where syntactic patterns alone are insufficient. Representative examples are shown below.

\begin{table}[h]
\centering
\caption{Security Guardrail Evaluation Results}
\label{tab:security}
\begin{tabular}{|l|l|c|}
\hline
\textbf{Attack Vector} & \textbf{Detection Layer} & \textbf{Latency} \\
\hline
SQL Injection (DROP) & Input Validation & $<$0.01s \\
Off-topic Query & Input Validation & 0.02s \\
DELETE Injection & Input Validation & $<$0.01s \\
UPDATE Injection & LLM Guardrail & 6.2s \\
UNION Injection & LLM Guardrail & 7.5s \\
Prompt Injection & LLM Guardrail & 6.3s \\
\hline
\end{tabular}
\end{table}

\subsubsection{Response Quality and Hallucination Detection}
Core functionality achieves an overall quality score of 4.52/5.0, with strong relevance (4.85) and near-perfect language quality (4.94) (Table~\ref{tab:quality}). The hallucination-free rate is 93.0\% for core tests (Table~\ref{tab:hallucination}), with hallucinations concentrated in enterprise dataset scenarios involving multilingual content and complex cross-platform joins.

\begin{table}[h]
\centering
\caption{LLM-as-a-Judge Response Quality (1--5 Likert Scale)}
\label{tab:quality}
\begin{tabular}{|l|c|c|}
\hline
\textbf{Dimension} & \textbf{Core ($n$=205)} & \textbf{All ($n$=260)} \\
\hline
Overall Score & 4.52 & 4.38 \\
Factual Accuracy & 4.28 & 4.05 \\
Relevance & 4.85 & 4.62 \\
Completeness & 4.30 & 4.08 \\
Actionability & 4.60 & 4.42 \\
Language Quality & 4.94 & 4.88 \\
\hline
\end{tabular}
\end{table}

\begin{table}[h]
\centering
\caption{Hallucination Detection Results}
\label{tab:hallucination}
\begin{tabular}{|l|c|c|}
\hline
\textbf{Metric} & \textbf{Core ($n$=205)} & \textbf{All ($n$=260)} \\
\hline
Hallucination-Free Rate & 93.0\% & 89.2\% \\
Factual Reliability Score & 4.34/5.0 & 4.12/5.0 \\
\hline
\end{tabular}
\end{table}

\subsubsection{Baseline Comparison}
The multi-agent system achieves a 22.6 percentage point accuracy improvement over the single-agent baseline (95.3\% vs.\ 72.7\%, Table~\ref{tab:baseline_accuracy}). Performance gaps widen with complexity: complex analytics (86.7\% vs.\ 50.0\%), multi-turn conversations (84.0\% vs.\ 36.0\%), and security (100\% vs.\ 52.5\%).

\begin{table}[h]
\centering
\caption{Functional Accuracy: Multi-Agent vs.\ Single-Agent Baseline}
\label{tab:baseline_accuracy}
\begin{tabular}{|l|c|c|}
\hline
\textbf{Category} & \textbf{Multi-Agent} & \textbf{Single-Agent} \\
\hline
Simple Aggregations & 64/65 (98.5\%) & 61/65 (93.8\%) \\
Filtered Aggregations & 35/35 (100\%) & 29/35 (82.9\%) \\
Time Series Analysis & 39/40 (97.5\%) & 34/40 (85.0\%) \\
Multi-Table JOINs & 33/35 (94.3\%) & 26/35 (74.3\%) \\
Complex Analytics & 26/30 (86.7\%) & 15/30 (50.0\%) \\
Edge Cases & 28/30 (93.3\%) & 23/30 (76.7\%) \\
Multi-Turn Conversations & 21/25 (84.0\%) & 9/25 (36.0\%) \\
Security Guardrails & 40/40 (100\%) & 21/40 (52.5\%) \\
\hline
\textbf{Total} & \textbf{286/300 (95.3\%)} & \textbf{218/300 (72.7\%)} \\
\hline
\end{tabular}
\end{table}

Table~\ref{tab:baseline_quality} compares quality and latency. The multi-agent pipeline achieves 20.2\% higher overall quality (4.52 vs.\ 3.76), with the largest gains in actionability (+27.1\%) and completeness (+26.5\%). The hallucination-free rate improves from 72.8\% to 93.0\%. The single-agent baseline achieves 36.1\% lower latency (15.2s vs.\ 23.8s), a trade-off unfavorable for enterprise deployments where correctness is paramount.

\begin{table}[h]
\centering
\caption{Quality and Latency: Multi-Agent vs.\ Single-Agent Baseline}
\label{tab:baseline_quality}
\begin{tabular}{|l|c|c|}
\hline
\textbf{Metric} & \textbf{Multi-Agent} & \textbf{Single-Agent} \\
\hline
Overall Quality & 4.52/5.0 & 3.76/5.0 \\
Factual Accuracy & 4.28/5.0 & 3.48/5.0 \\
Relevance & 4.85/5.0 & 4.25/5.0 \\
Completeness & 4.30/5.0 & 3.40/5.0 \\
Actionability & 4.60/5.0 & 3.62/5.0 \\
Language Quality & 4.94/5.0 & 4.08/5.0 \\
\hline
Hallucination-Free Rate & 93.0\% & 72.8\% \\
Mean Latency (s) & 23.8 & 15.2 \\
\hline
\end{tabular}
\end{table}

\subsubsection{Ablation Study}
Table~\ref{tab:ablation} reports quality metrics for ablated configurations on core tests ($n=205$). The Data Analysis Agent is the most impactful component (--1.14 quality points when removed), as the Report Aggregation Agent produces only surface-level descriptions without pre-extracted patterns. The Report Aggregation Agent contributes the next-largest improvement (--0.66 points), primarily through formatting consistency. Removing the Follow-up Questions and Chart Configuration agents has negligible impact (--0.02), confirming these contribute to engagement rather than core quality. Functional accuracy remains 95.3\% across all configurations.

\begin{table}[h]
\centering
\caption{Ablation Study: Impact of Agent Removal on Response Quality}
\label{tab:ablation}
\begin{tabular}{|l|c|c|c|}
\hline
\textbf{Configuration} & \textbf{Quality} & \textbf{Lang.} & \textbf{Halluc-Free} \\
\hline
Full pipeline (5 agents) & 4.52 & 4.94 & 93.0\% \\
w/o Follow-up + Chart & 4.50 & 4.94 & 93.0\% \\
w/o Report Aggregation & 3.86 & 3.42 & 90.4\% \\
w/o Data Analysis & 3.38 & 4.80 & 85.0\% \\
\hline
\end{tabular}
\end{table}

\subsubsection{Cross-Model Generalizability}
To validate that the observed improvements stem from architectural decomposition rather than model-specific capabilities, we replicate the full 300-test evaluation across four LLM backends (Table~\ref{tab:crossmodel}). GPT-4.1 achieves the highest accuracy (95.3\%) and quality (4.52), followed by Claude Sonnet 4 (93.0\%, 4.44). Gemini 2.5 Flash offers a latency advantage (17.4s, 27\% faster) at the cost of reduced accuracy (86.3\%). Critically, even Llama 3.1 70B (open-source, self-hosted) achieves 79.7\% accuracy within the multi-agent pipeline, surpassing the single-agent GPT-4.1 baseline (72.7\% per Table~\ref{tab:baseline_accuracy}). This confirms that architectural decomposition contributes more to output quality than model selection alone.

\begin{table}[h]
\centering
\caption{Cross-Model Evaluation on Full Test Suite (300 Tests)}
\label{tab:crossmodel}
\begin{tabular}{|l|c|c|c|c|}
\hline
\textbf{Model} & \textbf{Acc.} & \textbf{Quality} & \textbf{Halluc-Free} & \textbf{Latency} \\
\hline
GPT-4.1 & 95.3\% & 4.52 & 93.0\% & 23.8s \\
Claude Sonnet 4 & 93.0\% & 4.44 & 90.8\% & 28.2s \\
Gemini 2.5 Flash & 86.3\% & 4.02 & 82.1\% & 17.4s \\
Llama 3.1 70B & 79.7\% & 3.65 & 74.2\% & 35.8s \\
\hline
\end{tabular}
\end{table}

\subsubsection{Human Evaluation and Latency Optimization}
Three domain experts independently scored 100 randomly sampled analytical responses using the same five-dimension rubric. The mean human quality score of 4.38/5.0 correlates strongly with the LLM judge score of 4.52/5.0 (Pearson $r = 0.89$, $p < 0.001$; inter-annotator $\kappa = 0.82$). The LLM judge exhibits a modest positive bias of +0.14 points, concentrated in Actionability (+0.24), while Language Quality scores show near-perfect alignment (+0.03). This validates the LLM-as-a-Judge as a reliable proxy for human assessment.

We additionally evaluate selective parallelism by executing the Follow-up Questions and Chart Configuration agents concurrently with Report Aggregation. This reduces mean latency from 23.8s to 20.1s (15.5\% reduction) with no measurable change in quality (4.52) or accuracy (95.3\%), confirming the ablation finding that these agents operate independently.

\subsection{Discussion}
The system achieves 93.5\% SQL accuracy on analytical queries (243/260), competitive with SiriusBI (93--96\%) \cite{jiang2025siriusbi} and Intelli-Dispatch-SQL \cite{ni2025intellidispatch}. The 93.0\% hallucination-free rate validates the grounding pipeline where each agent operates on verified predecessor outputs, even when processing enterprise-scale schemas with multilingual content.

The baseline comparison provides direct empirical evidence for multi-agent decomposition: the 22.6 percentage point accuracy improvement and 20.2\% quality gain come at a modest 8.6-second latency cost per query. This latency overhead is attributable to sequential agent invocation; however, selective parallelism reduces the gap to 4.9 seconds (20.1s vs.\ 15.2s) while maintaining full output quality. The 20.1-second mean latency compares favorably with traditional analyst workflows requiring minutes to hours for equivalent analyses, and the single-agent baseline's failures concentrate precisely where specialized agents add value: complex SQL (50.0\%), multi-turn context (36.0\%), and security enforcement (52.5\%).

Token consumption profiling reveals a mean of approximately 30,800 tokens per query for the multi-agent pipeline versus 10,200 for the single-agent baseline. At GPT-4.1 pricing (\$2.00/M input, \$8.00/M output tokens), this translates to approximately \$0.08 and \$0.03 per query respectively, a 2.7$\times$ cost factor that yields 22.6 percentage points higher accuracy with 20.2 percentage points higher hallucination-free rate. The cross-model results (Table~\ref{tab:crossmodel}) further reveal that Gemini 2.5 Flash reduces per-query cost to approximately \$0.01 while still achieving 86.3\% accuracy, offering a cost-optimized alternative for latency-sensitive deployments.

The cross-model evaluation (Table~\ref{tab:crossmodel}) provides the strongest evidence that the pipeline architecture, not the underlying model, is the primary driver of quality. Even the open-source Llama 3.1 70B achieves 79.7\% multi-agent accuracy, exceeding the single-agent GPT-4.1 baseline by 7.0 percentage points. Quality scales predictably with model capability (Pearson $r = 0.96$ between model benchmark scores and pipeline quality), confirming that the architecture amplifies rather than replaces model intelligence. The human evaluation ($r = 0.89$, $\kappa = 0.82$, $n = 100$) validates the LLM-as-a-Judge methodology, with the +0.14-point positive bias falling within acceptable margins for large-scale automated assessment.

\subsection{Limitations}
We acknowledge several limitations. First, while the enterprise evaluation validates the system on a production competitive pricing dataset, it covers a single domain; generalizability to other verticals such as healthcare or logistics requires further study. Second, the human evaluation covers 100 of 260 analytical responses; broader human annotation would further strengthen validity. Third, cross-model evaluation spans four LLMs but does not yet cover all major model families or fine-tuned domain-specific variants. Fourth, the sequential architecture introduces inherent latency, partially mitigated by the selective parallelism results reported above.

\section{Conclusion}
This paper presented a multi-agent architecture for conversational business intelligence that achieves 95.3\% functional accuracy, 4.52/5.0 response quality, and 93.0\% hallucination-free rate across 300 test cases spanning synthetic and production enterprise datasets. Cross-model evaluation across four LLM backends confirms architectural generalizability, with even open-source Llama 3.1 70B (79.7\%) surpassing the single-agent GPT-4.1 baseline (72.7\%). Human expert evaluation of 100 analytical responses validates the LLM-as-a-Judge methodology ($r = 0.89$, $\kappa = 0.82$), and selective parallelism reduces latency by 15.5\% without quality loss. The defense-in-depth security architecture blocks all adversarial inputs, confirming readiness for multi-tenant enterprise deployment.

Future work will extend evaluation to additional enterprise verticals with longitudinal human-in-the-loop assessments, explore adaptive agent behavior through feedback-driven learning, investigate semantic caching for equivalent queries, and evaluate heterogeneous agent ensembles that assign cost-efficient routing models to auxiliary agents while reserving high-capability models for critical analytical tasks.

\end{document}